\documentclass[review]{elsarticle}
\graphicspath{ {./figures/} }
\usepackage{hyperref}
\usepackage{float}
\usepackage{verbatim} 
\usepackage{apalike}

\usepackage{amsmath} 
\usepackage{amssymb}
\usepackage{subcaption}

\restylefloat{figure}
\floatstyle{plaintop} 
\restylefloat{table}

\journal{Expert Systems with Applications}

\biboptions{authoryear}

\begin{document}
\begin{frontmatter}


\begin{titlepage}
\begin{center}
\vspace*{1cm}

\textbf{PACD-Net: Pseudo-Augmented Contrastive Distillation for Glycemic Control Estimation from SMBG}

\vspace{1.5cm}

Canyu Lei$^{a}$ (thq4up@virginia.edu), David Repaske$^{b}$ (DRR5C@uvahealth.org), Jianxin Xie$^{a}$ (hcf7fd@virginia.edu) \\

\hspace{10pt}

\begin{flushleft}
\small  
$^a$ University of Virginia, School of Data Science, Charlottesville, VA 22903, USA \\
$^b$ University of Virginia, Department of Pediatrics, Charlottesville, VA 22903, USA \\

\vspace{1cm}
\textbf{Corresponding author at: University of Virginia, School of Data Science, Charlottesville, VA 22903, USA} \\
Jianxin Xie \\
University of Virginia, School of Data Science, Charlottesville, VA 22903, USA \\
Tel: (850) 228-5895 \\
Email: hcf7fd@virginia.edu

\end{flushleft}        
\end{center}
\end{titlepage}

\title{PACD-Net: Pseudo-Augmented Contrastive Distillation for Glycemic Control Estimation from SMBG}

\author[label1]{Canyu Lei}
\ead{thq4up@virginia.edu}

\author[label2]{David Repaske}
\ead{DRR5C@uvahealth.org}

\author[label1]{Jianxin Xie\corref{cor1}}
\ead{hcf7fd@virginia.edu}

\cortext[cor1]{Corresponding author.}
\address[label1]{University of Virginia, School of Data Science, Charlottesville, VA 22903, USA}
\address[label2]{University of Virginia, Department of Pediatrics, Charlottesville, VA 22903, USA}

\begin{abstract}
Effective diabetes management requires continuous monitoring of glycemic levels. Clinically, glycemic control is assessed using metrics such as Time in Range (TIR), Time Below Range (TBR), and Time Above Range (TAR), typically derived from continuous glucose monitoring (CGM). However, many patients rely on self-monitoring of blood glucose (SMBG) due to the high cost and limited accessibility of CGM. Unlike CGM, SMBG provides sparse and irregular measurements, making accurate estimation of these metrics challenging. Conventional supervised learning approaches struggle under such sparsity, leading to poor generalization and unstable performance.
To address this, we propose PACD-Net, a self-supervised contrastive knowledge distillation framework for estimating glycemic control from SMBG. Pseudo-SMBG samples with richer temporal coverage are used as teacher signals to guide learning from sparse observations. In addition, multi-view contrastive learning enforces representation consistency across diverse sampling patterns. The model adopts a hybrid Swin Transformer–CNN backbone to capture temporal dependencies in sparse SMBG sequences. 
Experimental results demonstrate that PACD-Net consistently outperforms existing methods in estimating TAR, TIR, and TBR from real-world SMBG data, achieving improved accuracy as well as enhanced stability and generalization under extremely sparse observation settings. The proposed framework provides a practical tool for clinical SMBG interpretation and offers a generalizable approach for learning from sparse and irregularly sampled sensor data in broader applications.
\end{abstract}


\end{frontmatter}

\section{Introduction}
\label{introduction}

Diabetes is a chronic disease with a steadily increasing global prevalence, posing long-term and widespread threats to human health~\cite{khunti2023diabetes,lotfy2017chronic}. It is associated with a range of serious complications, including cardiovascular disease and microvascular complications such as diabetic retinopathy and nephropathy, which substantially increase the risks of disability and mortality~\cite{balaji2019complications,deshpande2008epidemiology,tomic2022burden}. These complications not only pose severe threats to individual patient outcomes but also place a sustained burden on global public health systems~\cite{susan2010global,arokiasamy2021global}. Extensive evidence has demonstrated that the development of diabetic complications is closely associated with the level of glycemic control~\cite{klein1998relation,skyler2004effects,stolar2010glycemic}. Consequently, close monitoring of glycemic control status is a critical component of diabetes prevention and management~\cite{vigersky2017role,cappon2019continuous,weinstock2020role}.

In clinical practice, glycemic control is commonly evaluated using time-based measures—including Time in Range (TIR), Time Above Range (TAR), and Time Below Range (TBR), collectively referred to as time-in-range (TR) metrics—which quantify the proportion of time that glucose levels fall within, above, or below a target range over a given period. These metrics provide a comprehensive assessment of glycemic control quality and variability, and have been shown to be closely associated with the risk of diabetes-related complications and treatment outcomes~\cite{aleppo2021clinical}.

TR metrics are primarily derived from two glucose monitoring modalities: continuous glucose monitoring (CGM) and self-monitoring of blood glucose (SMBG)~\cite{balaji2026use,zheng2020comparing}. CGM systems are wearable devices attached to the skin that measure glucose levels in interstitial fluid at regular intervals (e.g., every 5 minutes), enabling continuous glucose recording and providing accurate estimation of TR metrics~\cite{chobot2024exploring,maiorino2020effects}. However, the widespread adoption of CGM remains limited by its relatively high cost, restricted accessibility in resource-constrained populations, and challenges related to long-term wearability~\cite{rodbard2016continuous}. In contrast, SMBG relies on intermittent fingerstick measurements, offering a cost-effective alternative for glucose monitoring and is widely adopted as the glucose assessment tool in many low- and middle-income countries (LMICs)~\cite{ewen2025availability,huang2010cost}. 

Nonetheless, despite the widespread use of SMBG, glycemic control status (i.e., TR metrics) derived from SMBG suffers from limited accuracy due to the sparsity and irregularity of the measurements. Because SMBG measurements require finger-prick blood sampling, patients typically perform only a limited number of measurements per day~\cite{benjamin2002self}. Consequently, SMBG observations are inherently sparse, irregularly distributed over time, and highly heterogeneous across individuals in terms of sampling frequency and measurement habits, making reliable TR estimation difficult and complicating the clinical use of SMBG-based metrics for diabetes management~\cite{erbach2016interferences,hirsch2008self}.

Directly computing TR metrics from SMBG data using naive counting or interpolation therefore often introduces substantial systematic bias~\cite{avari2020differences, lei2025dpa}. Prior studies have shown that, when compared with CGM-derived ground truth, SMBG-based estimates tend to underestimate TIR while overestimating TAR and TBR, with such biases becoming more pronounced under shorter observation windows or lower sampling densities~\cite{aleppo2017replace}. Although a recent deep learning method~\cite{lei2025dpa} has been proposed to learn mappings from SMBG to TR metrics in a supervised manner using paired SMBG and CGM-derived labels, they often suffer from unstable training and limited generalization to diverse SMBG sampling patterns, with a tendency to overfit to the specific observation patterns present in the training data, due to the sparse and irregular nature of SMBG data. 

To address the abovementioned limitations, we propose a pseudo-augmented contrastive knowledge distillation framework, termed PACD-Net, to enable robust estimation of TR metrics from sparse SMBG data. Instead of learning a direct mapping from a single SMBG snapshot to TR metrics, PACD-Net captures sample-specific SMBG patterns by leveraging pseudo-augmented observations and contrastive distillation. Specifically, PACD-Net constructs multiple SMBG views with varying sampling patterns and sparsity levels, and enforces consistency in latent representations across these views with pseudo-augmented SMBG with enriched temporal coverage. This design mitigates the impact of limited supervision and improves generalization under highly sparse and irregular sampling conditions. The main contributions and significance of this work are summarized as follows:

\begin{itemize}
\item We develop a representation learning framework that learns sample-invariant embeddings across multiple SMBG views derived from the same underlying CGM profile. By leveraging contrastive learning, the model enforces consistency across these views while preserving discriminative features between different samples.

\item To address the sparsity of SMBG observations, we employ a student–teacher knowledge distillation framework that transfers information from pseudo-SMBG with richer temporal patterns (teacher) to a student encoder operating on highly sparse SMBG inputs, enabling more stable training and more robust, generalizable representations.

\item PACD-Net employs a modified Swin Transformer with convolutional residual blocks (CRBs) to capture both local and long-range dependencies in sparse SMBG sequences through efficient hierarchical self-attention.

\end{itemize}

The proposed approach provides a reliable and cost-effective alternative for glycemic assessment in settings where CGM is not widely available, with strong clinical applicability and high public health impact, particularly in low- and middle-income countries.

\section{Related Work}
\label{related_work}

Although SMBG provides limited temporal coverage and relatively sparse observations~\cite{schnell2015clinical}, it has historically been and remains a primary and widely used tool for monitoring glucose levels and evaluating glycemic control, despite the increasing availability of CGM systems~\cite{hirsch2008self, briggs2004self}. Because SMBG measurements are collected at only a few discrete time points each day, early research introduced a range of statistical and variability-based indicators to summarize glucose behavior using limited data. 

Among the earliest and most influential measures is the Mean Amplitude of Glycemic Excursions (MAGE), proposed by Service et al.~\cite{service1970mean}, to quantify major fluctuations in blood glucose levels. Basic statistical descriptors such as mean glucose and standard deviation (Mean and SD) were also widely adopted in major clinical studies, including the Diabetes Control and Complications Trial (DCCT)~\cite{effect1993diabetes} and the Kumamoto trial~\cite{ohkubo1995intensive}, as indicators of overall glycemic control. Subsequent work proposed additional indices designed to capture different aspects of glucose variability and risk. For instance, the J-index was introduced to combine information about average glucose and variability into a single metric~\cite{wojcicki1995j}, while the Low and High Blood Glucose Indices (LBGI and HBGI) were developed to quantify the risks associated with hypoglycemia and hyperglycemia~\cite{kovatchev1997symmetrization}. Other variability measures have also been proposed, including the coefficient of variation (CV\%)~\cite{monnier2008glycemic}, the Mean of Daily Differences (MODD)~\cite{molnar1972day}, which reflects day-to-day variability, and the Continuous Overall Net Glycemic Action (CONGA)~\cite{mcdonnell2005novel}, which evaluates glucose fluctuations over predefined time intervals. 
Furthermore, prior to the widespread adoption of CGM, several clinical practice guidelines incorporated metrics such as mean glucose, standard deviation (SD), and MAGE as surrogate indicators for evaluating glycemic control~\cite{american2005standards,american2006standards}. Although these indices provide useful summaries of glucose variability, they are inherently coarse qualitative measures that fail to capture temporal patterns and do not directly quantify critical clinical information such as time spent in target glycemic ranges. In contrast, TR metrics serve as more informative indicators, reflecting the overall quality of glycemic control and the progression of diabetes. 

With the rapid advancement of artificial intelligence, machine learning techniques have been increasingly applied to diabetes management and glucose prediction tasks~\cite{mujumdar2019diabetes, alam2024machine}. A growing body of work has explored the use of data-driven models to analyze glucose measurements and identify potential risks associated with abnormal glucose levels~\cite{nomura2021artificial, afsaneh2022recent}. Several studies have specifically investigated the feasibility of leveraging SMBG data for predictive modeling despite its sparse sampling characteristics. For instance, Sudharsan et al. developed machine learning models to predict hypoglycemic events in patients with type 2 diabetes using SMBG measurements~\cite{sudharsan2014hypoglycemia}. Oviedo et al. further proposed individualized prediction models to detect postprandial hypoglycemia based on capillary SMBG records collected from diabetic patients~\cite{oviedo2019minimizing}. In addition, Faruqui et al. developed deep learning approaches to forecast short-term glucose levels using routinely collected clinical data, including SMBG observations~\cite{faruqui2019development}. Woldaregay et al. applied machine learning techniques to classify blood glucose patterns and detect anomalies in type 1 diabetes datasets~\cite{woldaregay2019data}. These studies demonstrate that meaningful predictive insights can be obtained even from sparse SMBG data. However, most existing approaches primarily focus on event-level prediction or short-term glucose forecasting, while exploration of capturing complex temporal dependencies from sparse observations remains limited.

Recent advances in deep learning have introduced more powerful representation learning frameworks for modeling physiological time-series data in medical and healthcare applications~\cite{ruan2019representation,morid2023time}. 
Transformer-based architectures have demonstrated strong capabilities in capturing long-range dependencies and global contextual relationships, which makes them particularly suitable for modeling time-series data with complex temporal structures. 
For example, Zhu et al. proposed a Transformer-based framework for population-specific glucose prediction in diabetes care~\cite{zhu2024population}. Xu et al. introduced TransEHR, a self-supervised Transformer model designed for clinical time-series data~\cite{xu2023transehr}. In addition, Tipirneni and Reddy developed a self-supervised Transformer framework specifically designed for sparse and irregularly sampled multivariate clinical time-series~\cite{tipirneni2022self}. These studies highlight the potential of Transformer architectures for modeling complex temporal dependencies in healthcare data. 

In parallel, self-supervised learning has emerged as an effective paradigm for learning informative representations from large amounts of unlabeled data. Such approaches have been increasingly applied in medical and healthcare domains, where labeled data are often limited or expensive to obtain~\cite{liu2023self}.
For example, Lee et al. proposed a multi-view integrative attention-based deep representation learning framework for irregular clinical time-series data~\cite{lee2022multi}, while Kiyasseh et al. introduced a contrastive learning approach for ECG representation learning~\cite{kiyasseh2021clocs}. 
Existing self-supervised learning approaches construct multiple views through generic augmentations such as masking, cropping, or noise injection~\cite{he2020momentum, grill2020bootstrap}. While these strategies have shown success in learning invariant representations, they are not designed to reflect the irregular and behavior-driven sampling patterns inherent in sparse clinical measurements. To address this limitation, we propose PACD-Net, which purposefully generates multiple pseudo-SMBG views from the underlying CGM profiles, preserving realistic sparsity patterns while introducing diverse observation schemes. This design enables the model to learn invariant representations across sparse SMBG views, while simultaneously leveraging pseudo-SMBG with richer information to guide representation learning. As a result, PACD-Net effectively bridges sparse observations and latent glycemic dynamics for accurate estimation of time-in-range metrics.

\section{Methodology}

\subsection{Problem Statement}
CGM provides dense, stable, and high-frequency measurements of blood glucose and is widely regarded as a reliable reference for assessing glycemic control in both clinical and research practice~\cite{cappon2019continuous}. Accordingly, as defined in the Ambulatory Glucose Profile (AGP) report, glycemic control is commonly evaluated using a set of standardized CGM-based glycemic variability metrics, among which the most representative are \emph{Time in Range (TIR)}, \emph{Time Above Range (TAR)}, and \emph{Time Below Range (TBR)}~\cite{czupryniak2022ambulatory}. These metrics characterize the proportion of time that blood glucose values fall within different clinically relevant ranges, thereby providing an overall assessment of glycemic stability and the risks of hyperglycemia and hypoglycemia~\cite{johnson2019utilizing}.

\begin{figure}[htbp]
\centering
\includegraphics[width=0.5\linewidth]{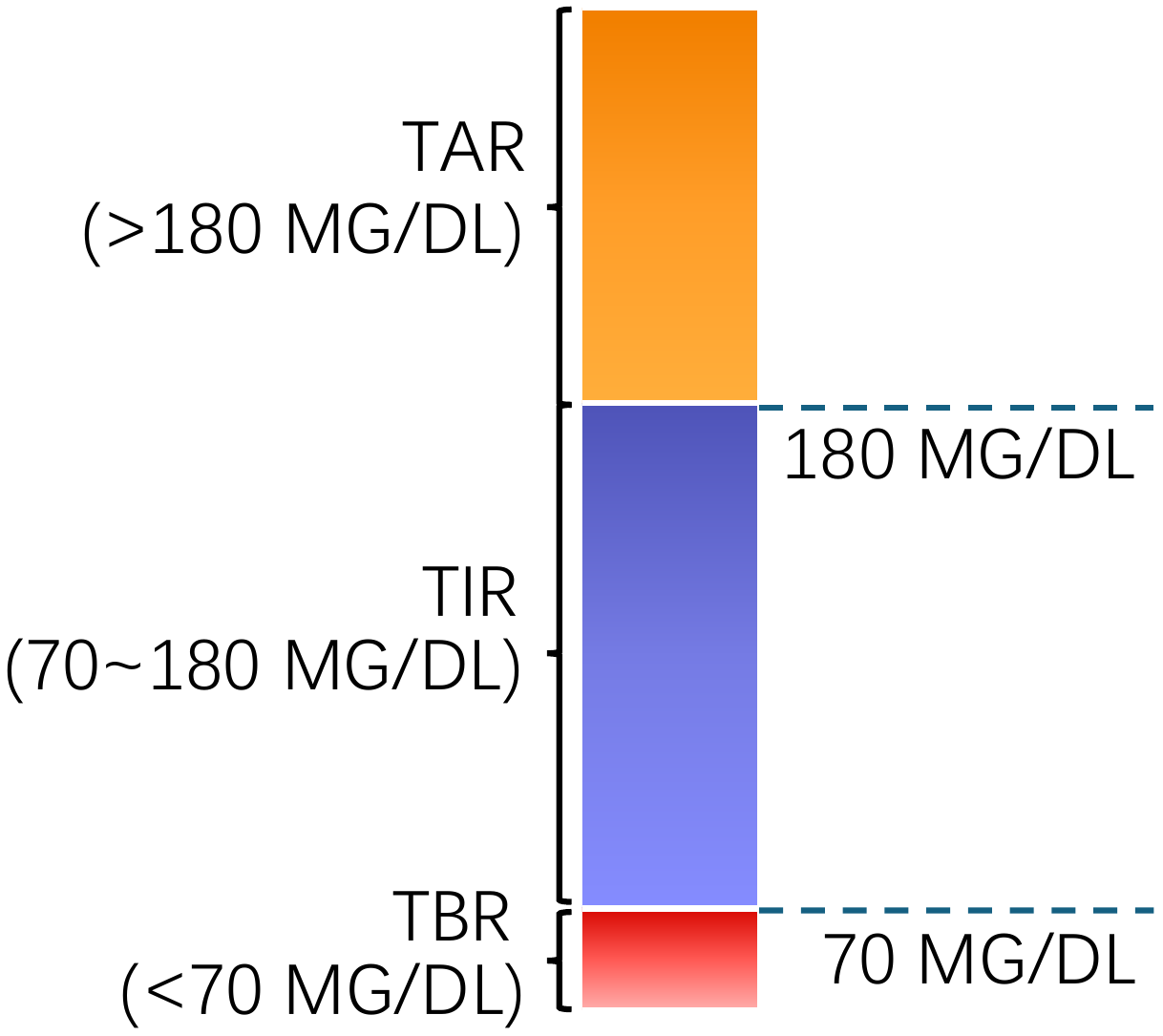}
\caption{AGP times-in-range “thermometer” (TAR/TIR/TBR).}
\label{fig:tr_description}
\end{figure}

In this work, glycemic control metrics are formally defined based on standard CGM thresholds. Specifically, \emph{TIR} is defined as the proportion of CGM readings within the target range of 70--180~mg/dL, \emph{TAR} as the proportion of readings exceeding the upper threshold $\tau_{\text{high}} = 180$~mg/dL, and \emph{TBR} as the proportion of readings below the lower threshold $\tau_{\text{low}} = 70$~mg/dL. Figure~\ref{fig:tr_description} illustrates these glucose ranges and their corresponding metric definitions.

Specifically, let $D$ denote the number of the observation days, and let $T$ denote the number of glucose measurements per day. According to the CGM sampling protocol, glucose values are recorded at a uniform interval of 5 minutes, resulting in $T = 288$ measurements per day. Let $g_{d,t}$ denote the CGM glucose value measured at day $d$ and time index $t$ for sample $\rho$, where $d = 1, \ldots, D$ and $t = 1, \ldots, T$. The glycemic control metrics TIR, TAR and TBR are formulated as the proportion of glucose measurements that fall within different clinically defined ranges over a $D$-day window, mathematically: 

\begin{align}
\label{Eq:cgm_to_tr}
\mathrm{TIR}_\rho &=\frac{1}{D T} \sum_{d=1}^D \sum_{t=1}^T \mathbf{1}\left(\tau_{\text {low}} \leq g_{d, t}^{(\rho)} \leq \tau_{\text {high }}\right)  \nonumber\\
\mathrm{TAR}_\rho &=\frac{1}{D T} \sum_{d=1}^D \sum_{t=1}^T \mathbf{1}\left(g_{d, t}^{(\rho)}>\tau_{\text {high }}\right) \nonumber\\
\mathrm{TBR}_\rho &=\frac{1}{D T} \sum_{d=1}^D \sum_{t=1}^T \mathbf{1}\left(g_{d, t}^{(\rho)}<\tau_{\text {low }}\right)
\end{align}

By construction, $\text{TAR}_\rho + \text{TIR}_\rho + \text{TBR}_\rho =1 $.

However, although CGM provides an accurate and reliable observational basis for computing AGP metrics, its use in real-world practice remains limited due to high device cost and accessibility~\cite{rodbard2021ambulatory}. In contrast, SMBG remains more widely used due to its affordability and its non-skin-attached mode of data collection. Nevertheless, SMBG observations are typically sparse, irregularly sampled, and temporally discontinuous, which hinders the accurate characterization of key glycemic control metrics. 

In the current investigation, we aim to extract glycemic control metrics (i.e., TR metrics) directly from SMBG data. To formalize the problem, the $\rho$-th SMBG sample spanning $D$ consecutive days can be expressed as

\begin{equation}
    M_{s,\rho} = \{s_{d,t,\rho} \mid d = 1,\ldots,D;\; t = 1,\ldots,T\} \in \mathbb{R}^{D \times T}
\label{Eq:M_s,rho}
\end{equation}

where $s_{d,t,\rho}$ denotes the glucose value observed for sample $\rho$ on day $d$ at time index $t$, and $s_{d,t,\rho} = 0$ if no SMBG measurement is available at $(d,t)$.

Each sample is represented as a pair $(M_{s,\rho}, \mathrm{TR}_\rho)$, where 
$ \mathrm{TR}_\rho = \{\mathrm{TAR}, \mathrm{TIR}, \mathrm{TBR}\}_\rho$ denotes the corresponding set of CGM-derived TR metrics, which is set as the TR ground truth.

The learning objective is to estimate a mapping function $f_\theta$ that predicts glycemic control metrics from sparse SMBG observations,
\begin{equation}
   f_\theta : M_{s,\rho} \longrightarrow \widehat{\mathrm{TR}}_\rho,
\end{equation}
where $f_\theta$ is parameterized by the proposed PACD-Net and is designed to deliver clinically meaningful and CGM-comparable glycemic control information using sparse SMBG.

\subsection{Multi-View Input Representation for Sparse SMBG Data}

In this study, we propose a self-supervised Swin Transformer-based pseudo-augmented contrastive distillation model (PACD-Net) for estimating long-term glycemic control metrics---TIR, TAR, and TBR---from sparse SMBG. The overall framework of the PACD-Net is illustrated in Fig.~\ref{fig:pacd}. Due to the irregular and sparse nature of SMBG observations, it is non-trivial to design SMBG input representations that preserve as much informative content as possible. Here, we construct an SMBG input composed of three complementary components, each designed to capture a specific characteristic:

\begin{figure*}[htbp]
\centering
\includegraphics[width=1.0\linewidth]{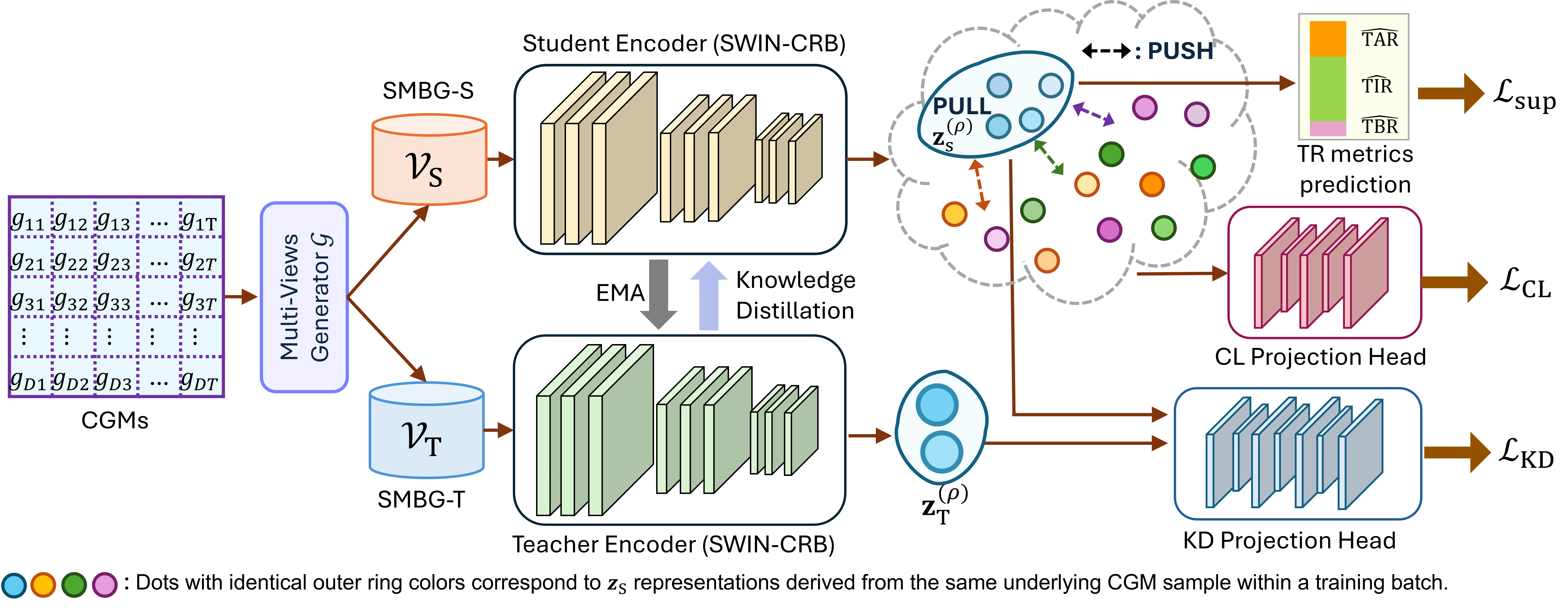}

\caption{Overview of the proposed PACD-Net architecture. }
\label{fig:pacd}
\end{figure*}

\begin{enumerate}
    \item \textbf{SMBG value matrix $M_{s,\rho}$:} 
    This matrix $M_{s,\rho}\in \mathbb{R}^{D\times T}$ encodes the SMBG measurements collected over a $D$-day observation window for $\rho$-th sample, as formally defined in Eq.~\ref{Eq:M_s,rho}.

    \item \textbf{Missing mask matrix $M_{m,\rho}$:} 
    To explicitly represent the sparse sampling pattern inherent to SMBG, we introduce a binary mask matrix that indicates whether a glucose measurement is observed at a temporal instance. This design allows the model to differentiate between unobserved measurements (missing entries) and valid glucose recordings, avoiding ambiguity that would otherwise arise from naively imputing missing entries with zeros. Formally, the mask matrix is defined as

    \begin{equation}
        M_{m,\rho} = \{m_{d,t,\rho} \mid d=1,\cdots,D;\; t=1,\cdots,T\} \in \mathbb{R}^{D \times T},
    \end{equation}
    where
    \begin{equation}
        m_{d,t,\rho} =
        \begin{cases}
            1, & \text{if the measurement for sample $\rho$ is absent at } (d,t), \\
            0, & \text{otherwise}.
        \end{cases}
    \end{equation}

    \item \textbf{Positional encoding matrix $M_{p,\rho}$:} 
    Since reshaping the one-dimensional SMBG sequence into a two-dimensional $D \times T$ grid removes explicit temporal ordering, an additional positional encoding is incorporated to preserve the underlying temporal structure. This encoding enables the model to recognize both intra-day temporal progression and inter-day periodicity within the 2D representation.

    Specifically, we adopt a two-dimensional sinusoidal positional encoding scheme~\cite{vaswani2017attention}. Temporal encodings are generated independently for the day dimension and the time-of-day dimension, and then combined to form a unified positional representation. Let $p$ denote the embedding index and $P$ the total encoding dimensionality. For the day index $i \in \{0,\ldots,D-1\}$, the encoding is defined as
    \begin{equation}
    \begin{split}
        \mathrm{PE}_{\mathrm{day}}[i, 2p]   &= \sin\left(\frac{i}{10000^{2p/P}}\right), \\
        \mathrm{PE}_{\mathrm{day}}[i, 2p+1] &= \cos\left(\frac{i}{10000^{2p/P}}\right),
    \end{split}
    \end{equation}
    Similarly, for the time-of-day index $j \in \{0,\ldots,T-1\}$,
    \begin{equation}
    \begin{split}
        \mathrm{PE}_{\mathrm{time}}[j, 2p]   &= \sin\left(\frac{j}{10000^{2p/P}}\right), \\
        \mathrm{PE}_{\mathrm{time}}[j, 2p+1] &= \cos\left(\frac{j}{10000^{2p/P}}\right).
    \end{split}
    \end{equation}

    The final positional encoding at each $(d,t)$ location is obtained by element-wise summation of the corresponding day and time encodings. The multi-channel encoding is further aggregated along the embedding dimension via summation, yielding a positional encoding matrix $M_{p,\rho} \in \mathbb{R}^{D \times T \times 1}$.
\end{enumerate}

The three input components are concatenated along the channel dimension to form the composite input tensor for one SMBG input:
\begin{equation}
    \Theta_{\rho} = \mathrm{Concat}\!\left(M_{s,\rho},\, M_{m,\rho},\, M_{p,\rho}\right) \in \mathbb{R}^{3 \times D \times T}.
\end{equation}

Note that, both SMBG and CGM data are required to train the neural network. Due to variability in patient behavior, SMBG measurements may be collected under various temporal sampling patterns, even when other conditions remain the same. This gives rise to multiple views of SMBG observations. For one particular sample, these different views, referred as pseudo-SMBG, are all derived from the same underlying CGM trajectory within an observation window spanning $D$ days. Consequently, despite differences in sampling patterns, all pseudo-SMBG views should correspond to the same underlying glycemic control status and share identical TR metrics. As such, we encode this invariance into our self-supervised learning framework.

In addition to the available real SMBG measurements (if any), we generate multiple pseudo-SMBG views from the CGM data to create diverse temporal sampling patterns. 
By repeatedly applying the pseudo-SMBG generator $\mathcal{G}$ to the 
$\rho$-th CGM sample, multiple pseudo-SMBG views can be generated. 
Here, we denote $\mathcal{G}(\cdot;\alpha,\mathcal{S})$ as the pseudo-SMBG generator parameterized by the observation ratio $\alpha$ and the sampling policy $\mathcal{S}$.

Two types of views are constructed for each sample: teacher views (SMBG-T) and student views (SMBG-S). SMBG-T views are sampled with a larger glucose observation ratio so that they capture richer glucose patterns, which serve as stable targets for representation alignment. In contrast, SMBG-S views adopt a higher level of observation sparsity to mimic the highly sparse and irregular sampling patterns that are commonly observed in real-world SMBG data. Formally, the SMBG-T views generated from $\rho$-th CGM sample are defined as
\begin{equation}
    \Theta_{\mathrm{T},\rho}^{(i)} =
    \mathcal{G}\!\left(
    \text{CGM}_\rho;\,\alpha_{\mathrm{T}},\mathcal{S}_{\mathrm{T}}
    \right),
    \quad i = 1,\ldots,N_\mathrm{T},
\end{equation}

where $N_\mathrm{T}$ is the total number of teacher SMBG views, and $\alpha_\mathrm{T}$ represents the teacher view observation ratio, defined as the proportion of the pseudo SMBG measurements relative to the total number of glucose readings in that CGM sample. $\mathcal{S}_{\mathrm{T}}$ denotes the sampling strategy for SMBG-T, and here, it corresponds to purely random temporal sampling, ensuring unbiased and diverse coverage. Similarly, the $j$-th student view for the $\rho$-th sample is defined:
\begin{equation}
    \Theta_{\mathrm{S},\rho}^{(j)} =
    \mathcal{G}\!\left(
    \text{CGM}_\rho;\,\alpha_{\mathrm{S}},\mathcal{S}_{\mathrm{S}}
    \right),
    \quad j = 1,\ldots,N_{S},
\end{equation}

where $N_\mathrm{S}$ denotes the total number of student SMBG views in that sample and $\alpha_\mathrm{S}$ represents the student view observation ratio. 
$\mathcal{S}_{\mathrm{S}}$ denotes the sampling strategy for generating SMBG-S. 
Specifically, we employ both random sampling and active point selection (APS)~\cite{lei2025dpa} with controlled stochasticity, enabling more realistic temporal sampling patterns that mimic patient SMBG acquisition behavior.

Note that $\alpha_{\mathrm{S}} < \alpha_{\mathrm{T}}$, which indicates that SMBG-S exhibits a higher data sparsity compared to SMBG-T.
Notably, pseudo-SMBG view is encoded using the same input engineering, including the SMBG value matrix, the missingness mask matrix, and the positional encoding matrix, ensuring a consistent representation for the downstream network processing. The complete multi-view input set for sample $\rho$ is therefore given by

\begin{equation}
\mathcal{V}_{\rho} = \mathcal{V}_{\mathrm{T},\rho} \cup \mathcal{V}_{\mathrm{S},\rho}.
\end{equation}
where $\mathcal{V}_{\mathrm{T},\rho} = \{ \Theta_{\mathrm{T},\rho}^{(i)}\}_{i=1}^{N_T}$ and $\mathcal{V}_{\mathrm{S},\rho} = \{ \Theta_{\mathrm{S},\rho}^{(j)}\}_{j=1}^{N_S}$. 
By exposing the model to multiple views with diverse sampling patterns and sparsity levels, the proposed multi-view strategy enables the learning of representations that are invariant to observation irregularity and robust to sparse SMBG measurements. This input engineering lays the foundation for subsequent self-supervised representation alignment objectives.

\subsection{PACD-Net Overview}

In real-world settings, SMBG sampling patterns vary substantially across patients, resulting in highly irregular and sparse observations~\cite{rodbard2007optimizing}. Approaches that rely solely on a \textit{single} SMBG snapshot for direct prediction of TR metrics are highly sensitive to sampling timestamps and sampling frequency, and thus struggle to capture glycemic control status in a stable and reliable manner.

To address these limitations, we propose PACD-Net, a pseudo-augmented contrastive knowledge distillation framework for accurately inferring time-in-range (TR) metrics from sparse SMBG data. The overall framework is illustrated in Fig.~\ref{fig:pacd}. The network backbone is built upon a modified Swin Transformer with convolutional skip connections, enabling effective modeling of both intraday and interday temporal dependencies in blood glucose dynamics. Since the primary objective of PACD-Net is to estimate TR metrics, the main supervision is provided by aligning the predicted TR metrics from multiple views with the ground truth TR metrics derived from CGM data.

To mitigate the impact of the irregular and behavior-driven sampling patterns of SMBG data, we develop a self-supervised learning framework that leverages multiple pseudo-SMBG views generated from the corresponding CGM sample. If the pseudo views originate from the same CGM sample, their latent representations are expected to be consistent. To enforce this property, we introduce a contrastive learning objective that encourages representations of views from the same CGM sample to be close in the latent space, while pushing apart representations from different samples.

Another key challenge lies in the extreme sparsity of SMBG data, which provides limited information about the underlying glycemic control status and may lead to unstable neural network training and TR metric predictions. To mitigate this issue and improve training stability, we introduce a knowledge distillation mechanism. Specifically, SMBG-T views, which retain richer observational information, are used to guide the learning of SMBG-S views through a teacher–student framework. The teacher network, trained on denser views, provides more reliable supervision, thereby improving the robustness and consistency of the student network under sparse observation settings. The detailed model structures are introduced in the following subsections.

\subsection{Backbone Architecture: Swin Transformer with Convolutional Skip Connections}

In PACD-Net, both the student and teacher encoders share the same neural network architecture, which serves as the backbone for extracting latent representations from the structured SMBG input $\Theta_{\rho}$. Although the SMBG data are engineered into image-like input $\Theta_{\rho}$, the underlying temporal dependencies remain highly irregular and non-stationary on both intraday and interday scales. Conventional neural networks (CNNs), which rely on fixed local receptive fields and stationarity assumptions, are therefore limited in capturing such sparse temporal patterns. To address this, we adopt a Swin Transformer as the core backbone, which introduces hierarchical and shifted-window self-attention to enable adaptive and progressive modeling of non-local dependencies across temporal regions. However, local continuity and short-term glucose patterns remain critical and may not be fully preserved by attention alone. To this end, convolutional skip connections are incorporated to enhance local feature extraction and stabilize representation learning. This hybrid design (named Swin-CRB) effectively combines the strengths of content-adaptive global modeling and locality-preserving convolution, resulting in a more robust and expressive representation of glycemic dynamics under sparse and irregular sampling conditions.

\begin{figure*}[htbp]
\centering
\includegraphics[width=1.0\linewidth]{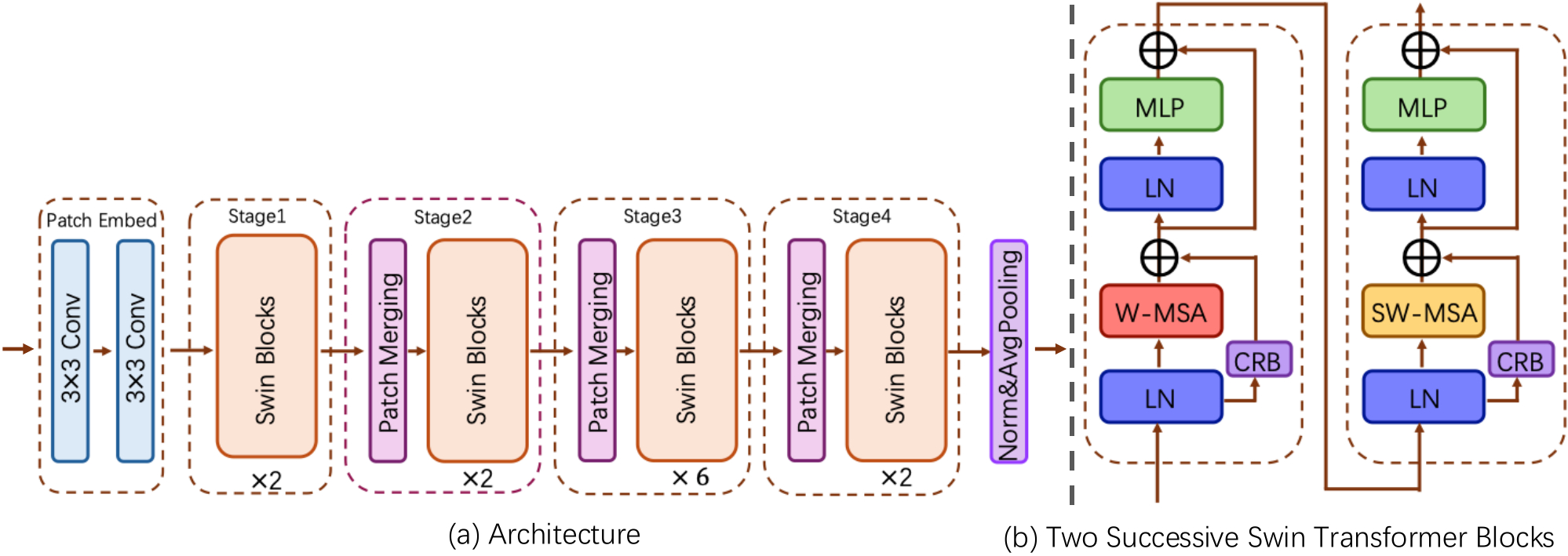}
\caption{The Architecture of the Swin-CRB backbone in PACD-Net.}
\label{fig:swin-crb}
\end{figure*}

The architecture of Swin-CRB is illustrated in Fig.~\ref{fig:swin-crb}(a). The input $\Theta_{\rho}$ is first processed by a convolutional patch embedding module composed of two consecutive $3 \times 3$ convolutional layers. This module progressively extracts local features while reducing the spatial resolution through strided convolutions, resulting in a lower-resolution feature map. The resulting feature map is then flattened into a sequence of patch tokens for subsequent Transformer encoding.

The resulting embeddings are then processed by a hierarchy of $L$ Swin Transformer stages. The first stage consists solely of a stack of Swin Transformer blocks operating at the original resolution. For subsequent stages, each stage begins with a patch merging module followed by multiple Swin Transformer blocks. The patch merging operation aggregates neighboring tokens and applies a linear projection to progressively reduce spatial resolution while increasing channel dimensionality, thereby constructing hierarchical multi-scale representations. After the final stage, layer normalization and global average pooling are applied to obtain a global feature representation for downstream tasks.

The basic structure of a Swin Transformer block follows that of a standard Transformer block~\cite{vaswani2017attention}, consisting of a self-attention submodule and a multi-layer perceptron (MLP) submodule connected in sequence. Layer normalization (LayerNorm, LN) is applied before each submodule to stabilize the training process. The key difference lies in the self-attention mechanism: instead of employing global multi-head self-attention (MSA) as in the standard Transformer, Swin Transformer adopts window-based self-attention to reduce computational complexity while preserving local modeling capability. Specifically, the self-attention submodule in a Swin Transformer block is implemented using either window-based multi-head self-attention (W-MSA) or shifted window-based multi-head self-attention (SW-MSA)~\cite{liu2021swin}. 

To further enhance the representation of local continuity patterns, a parallel convolutional residual branch (Convolutional Residual Branch, CRB) is introduced within the window-based self-attention submodule, as shown in Fig.~\ref{fig:swin-crb}(b). This branch consists of lightweight convolutional operations to extract local contextual features, which are fused with the window-based self-attention output at the residual level.
In the subsequent multilayer perceptron (MLP) submodule, the model follows the standard Transformer design, consisting of two fully connected layers with a non-linear activation in between with residual connections.

Within each Swin-CRB stage, multiple block pairs are employed, each comprising two consecutive blocks that alternately employ window-based multi-head self-attention (W-MSA) and shifted window-based multi-head self-attention (SW-MSA).
Let $l$ denote the index of the stage, and $b$ denote the index of the block pair within the $l$-th stage. 
The input feature map to the $b$-th block pair in stage $l$ is denoted as $\mathbf{z}^{l,b-1}$.
In the W-MSA block, the feature grid $\mathbf{z}^{l,b-1} \in \mathbb{R}^{H \times W \times C}$ is partitioned into non-overlapping windows of fixed size, and multi-head self-attention is computed independently within each window. In the subsequent SW-MSA, periodic window shifting is introduced to enable cross-window information interaction.
Mathematically, the Swin-CRB block pair at stage $l$ and index $b$ can be formulated as:

\begin{equation}
\begin{aligned}
\hat{\mathbf{z}}^{l,b}_{1} &=
\mathrm{W\text{-}MSA}\!\left(\mathrm{LN}(\mathbf{z}^{l,b-1})\right)
+ \mathrm{CRB}(\mathbf{z}^{l,b-1}), \\
\mathbf{z}^{l,b}_{1} &=
\mathrm{MLP}\!\left(\mathrm{LN}(\hat{\mathbf{z}}^{l,b}_{1})\right)
+ \hat{\mathbf{z}}^{l,b}_{1}, \\
\hat{\mathbf{z}}^{l,b}_{2} &=
\mathrm{SW\text{-}MSA}\!\left(\mathrm{LN}(\mathbf{z}^{l,b}_{1})\right)
+ \mathrm{CRB}(\mathbf{z}^{l,b}_{1}), \\
\mathbf{z}^{l,b} &=
\mathrm{MLP}\!\left(\mathrm{LN}(\hat{\mathbf{z}}^{l,b}_{2})\right)
+ \hat{\mathbf{z}}^{l,b}_{2}.
\end{aligned}
\end{equation}

where $\mathbf{z}^{l,b}$ denotes the output feature of the $b$-th block pair in stage $l$, and $\hat{\mathbf{z}}^{l,b}_{1}$ and $\hat{\mathbf{z}}^{l,b}_{2}$ denote the intermediate outputs of the W-MSA and SW-MSA blocks, respectively. 
W-MSA(·) and SW-MSA(·) denote window-based and shifted window-based self-attention operations, respectively. CRB(·) denotes the convolutional residual branch applied in parallel to the self-attention module.

\subsection{Knowledge distillation configuration}
\label{sec:KD}

SMBG observations are inherently sparse, which makes supervised training that directly maps SMBG inputs to TR metrics unstable and prone to unreliable predictions. The limited observational coverage introduces excessive degrees of freedom in the prediction space, making it difficult for the model to consistently capture the underlying glycemic control status.

To address this limitation, we introduce a self-supervised knowledge distillation (KD) framework~\cite{caron2021emerging}. In this design, the student encoder processes SMBG-S views that reflect realistic sparse sampling patterns, while the teacher encoder operates on SMBG-T views that retain richer observational information from the same glucose monitoring period. The student is trained to produce representations consistent with those of the teacher, thereby inheriting more informative and stable glycemic patterns from the teacher encoder. This mechanism not only stabilizes training but also enhances the representation learning capability of the student encoder under sparse observation settings.

Both the teacher and student encoders adopt the same Swin-CRB backbone architecture to ensure structural consistency in representation learning. The two networks differ only in their parameter update strategies. Specifically, the student encoder is optimized via standard backpropagation, while the teacher encoder is updated as an exponential moving average (EMA) of the student parameters. During training, the teacher network remains frozen with respect to gradient updates, providing a stable and slowly evolving target for the student. This design enables the teacher to capture more consistent and noise-robust representations, which in turn guide the student toward improved learning under sparse observation conditions.

Specifically, for $\rho$-th sample with ground truth CGM, the multi-view generator $\mathcal{G}$ produces a set of SMBG-T views $\mathcal{V}_{\mathrm{T},\rho}$ and a set of SMBG-S views $\mathcal{V}_{\mathrm{S},\rho}$. Let $v_\mathrm{T} \in \mathcal{V}_{\mathrm{T},\rho}$ and $v_\mathrm{S} \in \mathcal{V}_{\mathrm{S},\rho}$ denote arbitrary SMBG-T and SMBG-S views, respectively, for $\rho$-th sample. Each view is independently processed by the corresponding encoder and the shared KD projection head. The corresponding latent representations are defined as
\begin{equation}
    \mathbf{z}^{(\rho)}_{\mathrm{T}}(v_{\mathrm{T}}) = h\!\left(f_{\mathrm{T}}(v_{\mathrm{T}})\right), \quad
    \mathbf{z}^{(\rho)}_{\mathrm{S}}(v_{\mathrm{S}}) = h\!\left(f_{\mathrm{S}}(v_{\mathrm{S}})\right),
\end{equation}
where $f_{\mathrm{T}}(\cdot)$ and $f_{\mathrm{S}}(\cdot)$ denote the teacher and student encoders, and $h(\cdot)$ is the projection head.

The teacher outputs are centered and sharpened to form a target probability distribution over the representation space, which serve as soft targets for the student network:
\begin{equation}
    \mathbf{p}^{(\rho)}_{\mathrm{T}}(v_{\mathrm{T}}) = 
    \mathrm{softmax}\!\left(
    \frac{\mathbf{z}^{(\rho)}_{\mathrm{T}}(v_{\mathrm{T}}) - \mathbf{c}}{\tau_{\mathrm{T}}}
    \right),
\end{equation}
where $\tau_{\mathrm{T}}$ is the teacher temperature that controls the sharpness of the output distribution and $\mathbf{c}$ is a centering vector computed as the running average of the teacher outputs across samples, which is updated using a momentum-based exponential moving average (EMA) of the batch-wise mean teacher outputs:

\begin{equation}
    \mathbf{c} \leftarrow m\,\mathbf{c}
+ (1 - m)\,\frac{1}{B |\mathcal{V}_{\mathrm{T},\rho}|} \sum_{\rho=1}^{B} \sum_{v_{\mathrm{T}} \in \mathcal{V}_{\mathrm{T},\rho}} \mathbf{z}^{(\rho)}_{\mathrm{T}}(v_{\mathrm{T}})
\label{eq:center_update}
\end{equation}
where $m$ denotes the center momentum coefficient, $B$ is the batch size, and $\mathbf{z}^{(\rho)}_{\mathrm{T}}(v_{\mathrm{T}}) \in \mathbb{R}^D$ denotes the projection head output of the teacher encoder for a single view $v_{\mathrm{T}}$ of the $\rho$-th sample.

The student probability distribution is computed using a different temperature $\tau_{\mathrm{S}}$:
\begin{equation}
    \mathbf{p}^{(\rho)}_{\mathrm{S}}(v_{\mathrm{S}}) = 
    \mathrm{softmax}\!\left(
    \frac{\mathbf{z}^{(\rho)}_{\mathrm{S}}(v_{\mathrm{S}})}{\tau_{\mathrm{S}}}
    \right).
\end{equation}

The KD loss is designed to align the student representations with the teacher-generated target distributions. Specifically, for each SMBG-S view, the student output is first projected into a probability distribution, and then encouraged to match the corresponding teacher distribution obtained from richer SMBG-T views. This is achieved by minimizing the cross-entropy between the teacher and student distributions:
\begin{equation}
    \mathcal{L}_{\mathrm{KD}}
    =
    -\frac{1}{|\mathcal{V}_{\mathrm{T},\rho}|\,|\mathcal{V}_{\mathrm{S},\rho}|}
    \sum_{v_{\mathrm{T}} \in \mathcal{V}_{\mathrm{T},\rho}}
    \sum_{v_{\mathrm{S}} \in \mathcal{V}_{\mathrm{S},\rho}}
    \mathbf{p}^{(\rho)}_{\mathrm{T}}(v_{\mathrm{T}})^{\top}
    \log \mathbf{p}^{(\rho)}_{\mathrm{S}}(v_{\mathrm{S}}).
\label{eq:KD_loss}
\end{equation}

In this KD objective, the teacher network serves as a fixed target (stop-gradient), and only the student network is optimized via backpropagation.
This loss encourages the backbone to learn representations that are invariant to sampling sparsity and view-specific perturbations, thereby capturing the underlying glycemic dynamics even when the sampling frequency is low.

\subsection{Multi-View Contrastive Learning for SMBG-S Representations}

While the KD framework addresses the sparsity of SMBG data by enabling the student to learn richer representations from SMBG-T views with higher observation density, the irregularity arising from diverse and stochastic sampling behaviors remains unaddressed. Although the underlying blood glucose trajectory is unique for each sample (i.e., the CGM is fixed), the corresponding SMBG observations may exhibit substantial variation due to different temporal collection patterns.

To mitigate this uncertainty, we introduce a multi-view contrastive learning scheme that enforces discriminative and consistent representations across different SMBG-S views of the same sample. Specifically, this contrastive objective encourages views derived from the same CGM trajectory to be close in the embedding space, while pushing apart views from different samples. It is worth noting that this contrastive learning mechanism operates exclusively on the student embedding space, thereby enhancing the robustness of representations learned from sparse and irregular SMBG observations.

Specifically, for the $\rho$-th blood glucose sample, $N_S$ student views are constructed using the generator $\mathcal{G}$. SMBG-S views are independently processed by the student encoder to obtain student feature representations $
\{\mathbf{z}_{\mathrm{S},i}^{(\rho)}\}_{i=1,\ldots,N_{\mathrm{S}},\; \rho=1,\ldots,B}
$,
where the subscript $i$ indexes the $i$-th SMBG-S view of the $\rho$-th sample within a minibatch of size $B$. 
For a given anchor embedding $\mathbf{z}_{\mathrm{S},i}^{(\rho)}$, the positive pairs are defined as embeddings corresponding to other SMBG-S views of the same sample, i.e., $\{\mathbf{z}_{\mathrm{S},j}^{(\rho)} \mid j \neq i\}$. In contrast, the negative pairs are formed by embeddings from different samples within the batch, i.e., $\{\mathbf{z}_{\mathrm{S},k}^{(\rho')} \mid \rho' \neq \rho,\; k=1,\ldots,N_{\mathrm{S}}\}$.
This formulation encourages embeddings derived from the same underlying CGM trajectory to be close in the latent space, while pushing apart those originating from different samples.

Based on the above formulation, we adopt a contrastive learning objective to enforce similarity between positive pairs while distinguishing negative pairs. For a given anchor embedding $\mathbf{z}_{\mathrm{S},i}^{(\rho)}$, the contrastive loss is defined using an InfoNCE formulation: 

\begin{equation}
\begin{aligned}
&\mathcal{L}_{\mathrm{CL}} = \\
&- \sum_{\rho=1}^{B} \sum_{i=1}^{N_{\mathrm{S}}}
\log 
\frac{
\sum\limits_{j \neq i} \exp\!\left( \mathrm{sim}\!\left(\mathbf{z}_{\mathrm{S},i}^{(\rho)}, \mathbf{z}_{\mathrm{S},j}^{(\rho)}\right) / \tau_\text{CL} \right)
}{
\sum\limits_{\rho'=1}^{B} \sum\limits_{k=1}^{N_{\mathrm{S}}}
\mathbf{1}_{(\rho',k) \neq (\rho,i)} \;
\exp\!\left( \mathrm{sim}\!\left(\mathbf{z}_{\mathrm{S},i}^{(\rho)}, \mathbf{z}_{\mathrm{S},k}^{(\rho')}\right) / \tau_{\text{CL}} \right)
}
\end{aligned}
\end{equation}
where $\mathrm{sim}(\cdot,\cdot)$ denotes cosine similarity and $\tau_{\text{CL}}$ is a temperature parameter.

This objective promotes instance-level discrimination by pulling together embeddings from the same sample and pushing apart those from different samples, thereby enhancing robustness to SMBG sampling variability.

\subsection{Supervised TR Metric Prediction from Sparse SMBG}

Finally, the predicted TR metrics are required to be consistent with the ground truth values derived from CGM data. To ensure that the student feature embeddings effectively capture the underlying glycemic control status, we introduce a supervised loss that enforces alignment between the predicted TR metrics and their corresponding ground truth values, i.e., $\mathrm{TIR}, \mathrm{TAR}, \mathrm{TBR}$.

It is worth noting that the teacher encoder does not receive any supervised signals during training. Instead, its parameters are updated as an exponential moving average (EMA) of the student encoder parameters, as described in Section~\ref{sec:KD}.

Specifically, for each sample $\rho$ and the corresponding SMBG-S views $v_{\mathrm{S}} \in \mathcal{V}_{\mathrm{S},\rho}$, the student encoder produces a latent representation $\mathbf{z}_{\mathrm{S}}^{(\rho)}$, which is further mapped to TR metric predictions through a regression head $g(\cdot)$. The predicted TR metrics are denoted as $\widehat{\mathbf{y}}^{(\rho)}(v_{\mathrm{S}}) = g\!\left(\mathbf{z}_{\mathrm{S}}^{(\rho)}\right)$, where $\widehat{\mathbf{y}}^{(\rho)} = [\widehat{\mathrm{TIR}}, \widehat{\mathrm{TAR}}, \widehat{\mathrm{TBR}}]$.

The supervised learning objective enforces consistency between the predicted TR metrics and the ground truth values $\mathbf{y}^{(\rho)}$ derived from CGM data. This is achieved by minimizing a regression loss across all samples and views, defined as

\begin{equation}
\mathcal{L}_{\mathrm{sup}} = 
\frac{1}{N} \sum_{\rho=1}^{N} 
\frac{1}{|\mathcal{V}_{\mathrm{S},\rho}|} \sum_{v_{\mathrm{S}} \in \mathcal{V}_{\mathrm{S},\rho}}
\left\| \widehat{\mathbf{y}}^{(\rho)}(v_{\mathrm{S}}) - \mathbf{y}^{(\rho)} \right\|_2^2.
\end{equation}

where $N$ denotes the total number of samples. This formulation ensures that all SMBG-S views associated with the same sample are encouraged to produce consistent and accurate TR metric predictions.

The overall training objective of PACD-Net is formulated as a weighted combination of the supervised loss, the knowledge distillation (KD) loss, and the contrastive learning (CL) loss. Specifically, the total loss is defined as
\begin{equation}
\mathcal{L}_{\mathrm{total}} = 
\lambda_{\mathrm{sup}} \mathcal{L}_{\mathrm{sup}} +
\lambda_{\mathrm{KD}} \mathcal{L}_{\mathrm{KD}} +
\lambda_{\mathrm{CL}} \mathcal{L}_{\mathrm{CL}},
\end{equation}
where $\lambda_{\mathrm{sup}}, \lambda_{\mathrm{KD}}, \lambda_{\mathrm{CL}}$ are weighting coefficients that balance the contributions of each component.
The supervised loss ensures accurate prediction of TR metrics, the KD loss transfers stable and informative knowledge from richer SMBG-T views to sparse SMBG-S views, and the contrastive loss enforces sampling-invariant and discriminative representations across different views.

\section{EXPERIMENTAL DESIGN AND RESULTS}

\subsection{Datasets}
\label{sec:datasets}

This study leverages publicly available datasets obtained from the Jaeb Center for Public Health Data Repository. 
Among the available datasets, the \texttt{repbg} cohort from the REPLACE-BG study is of particular importance, as it contains paired CGM and SMBG data concurrently collected under real-world conditions. 
This unique characteristic enables two key advantages. First, the SMBG records reflect realistic patient sampling behaviors, which serve as a guideline for sampling pseudo SMBG-S from CGM considering the real behavioral patterns. Second, the corresponding CGM measurements provide reliable ground-truth TR metrics, enabling quantitative evaluation of the proposed model using real-world data.
The training set used in this study is constructed by combining all non-\texttt{repbg} CGM datasets with a subset of samples from the \texttt{repbg} dataset. 
The remaining \texttt{repbg} samples are reserved \textit{exclusively} for validation.
To ensure a strict subject-level separation, there is no patient overlap between the training and validation sets, and all reported results are obtained on the validation set.

Because most CGM datasets do not include simultaneously collected SMBG measurements, synthetic SMBG observations are generated for the training set by subsampling CGM time series.
Specifically, SMBG sampling times are inferred using either (i) random subsampling or (ii) an active point selection strategy~\cite{lei2025dpa}. 
Ground-truth TR metrics are computed directly from CGM data for both training and validation samples, as demonstrated in Eq.~\ref{Eq:cgm_to_tr}.
Validation is performed solely on the \texttt{repbg} dataset, where each real-world SMBG sample is paired with its corresponding CGM trace.
This evaluation setting closely reflects clinical practice and provides a realistic assessment of model robustness and applicability under real SMBG acquisition conditions.

\subsection{Evaluation Metrics}

Model performance is assessed using three standard regression metrics: root mean squared error (RMSE), coefficient of determination ($R^2$), and mean absolute error (MAE). 
All evaluations are conducted on the held-out validation set, which is composed of real-world SMBG samples and therefore provides a realistic benchmark for assessing generalization under practical SMBG sampling patterns.
Let $\text{TR}^{(j)}_\rho$ denote the ground-truth value of the $j$-th TR metric ($j \in \{\mathrm{TAR}, \mathrm{TIR}, \mathrm{TBR}\}$) for the $\rho$-th sample, and $\widehat{\text{TR}}^{(j)}_\rho$ represent the corresponding model prediction. 
The evaluation metrics are defined as follows:

\begin{align}
\mathrm{RMSE}^{(j)} &= \sqrt{\frac{1}{N_\text{v}} \sum_{\rho=1}^{N_\text{v}} \left(\text{TR}^{(j)}_\rho - \widehat{\text{TR}}^{(j)}_{\rho}\right)^2 \nonumber}\\
\mathrm{R}^{2(j)} &= 1 - \frac{\sum_{\rho=1}^{N_\text{v}} \left(\text{TR}^{(j)}_\rho - \widehat{\text{TR}}^{(j)}_\rho\right)^2}{\sum_{\rho=1}^{N_\text{v}} \left(\text{TR}^{(j)}_\rho - \overline{\text{TR}}^{(j)}\right)^2}\nonumber\\
\mathrm{MAE}^{(j)}  &= \frac{1}{N_\text{v}} \sum_{\rho=1}^{N_\text{v}} \left|\text{TR}^{(j)}_\rho - \widehat{\text{TR}}^{(j)}_{\rho}\right| \nonumber\\
\end{align}

where $N_\text{v}$ denotes the total number of samples in the validation set, and $\overline{\text{TR}}^{(j)}$ denotes the mean of the ground-truth values for the $j$-th target metric. 
To provide an overall performance summary, the aggregate RMSE and $R^2$ scores are computed as the average across the three target metrics:
\begin{align}
\mathrm{RMSE} &= \frac{1}{3}\sum_{j\in \mathrm{\{TAR, TIR, TBR\}}} \mathrm{RMSE}^{(j)} \nonumber\\
\mathrm{R}^2 &= \frac{1}{3}\sum_{j\in \mathrm{\{TAR, TIR, TBR\}}} \mathrm{R}^{2(j)}\nonumber\\
\end{align}

Lower values of RMSE and MAE indicate more accurate predictions, while $R^2$ values closer to 1 reflect a stronger agreement between predicted and ground-truth glycemic metrics.

\subsection{Effect of Sampling Sparsity in SMBG-T}

This subsection aims to systematically examine the effect of teacher-view observation sparsity on model prediction performance under a fixed student-view effective observation ratio $\alpha_\text{S}$.
The value of $\alpha_\text{S}$ is fixed at 3\% to reflect realistic clinical settings. 
Under this practical constraint, we vary the teacher-view effective observation ratio $\alpha_\text{T}$ to investigate how the amount of observational information provided by teacher views influences representation learning and downstream prediction accuracy.

\begin{table*}[htbp]
\caption{Effect of teacher-view effective observation ratio ($\alpha_\text{T}$) on model performance under a fixed student-view constraint.
The student-view effective observation ratio ($\alpha_\text{S}$) is fixed at 3\%, while $\alpha_\text{T}$ is varied.}
\centering
\scriptsize
\label{tab:teacher_rate_sweep}
\renewcommand{\arraystretch}{1.5}
\setlength{\tabcolsep}{6pt}
\begin{tabular}{|c|ccccccc|}
\hline
\text{Group} & \text{$\alpha_\text{T}$} & \text{$\alpha_\text{S}$} & \text{Overall RMSE} & \text{Overall $\text{R}^2$} & \text{TAR MAE} & \text{TIR MAE} & \text{TBR MAE}\\
\hline
T1            & 10\% & 3\% & 0.0589 & 0.6237 & 0.0559 & 0.0524 & 0.0168 \\ \hline
T2            & 30\% & 3\% & 0.0555 & 0.6518 & 0.0527 & 0.0504 & 0.0165 \\ \hline
T3            & 50\% & 3\% & 0.0533 & 0.6771 & 0.0500 & 0.0474 & 0.0162 \\ \hline
T4            & 70\% & 3\% & 0.0599 & 0.6346 & 0.0565 & 0.0532 & 0.0165 \\
\hline
\end{tabular}
\end{table*}

The experimental results are summarized in Table~\ref{tab:teacher_rate_sweep}. 
Under a fixed student-view effective observation ratio of $\alpha_{\text{S}} = 3\%$, 
the teacher-view effective observation ratio $\alpha_{\text{T}}$ has a significant impact on model prediction performance. 
Specifically, as $\alpha_{\text{T}}$ increases from 10\% to 30\% and further to 50\%, 
overall model performance exhibits a consistent improvement, characterized by a gradual decrease in RMSE and a corresponding increase in $R^2$, 
reaching its optimum at $\alpha_{\text{T}} = 50\%$. 
At this setting, the overall RMSE attains its minimum value (0.0533) and the overall $R^2$ reaches its maximum (0.6771), while the MAE values for TAR, TIR, and TBR are also the lowest among all configurations. 
These results indicate that when $\alpha_{\text{T}}$ is low, the limited amount of observational information retained in teacher views is insufficient to form stable and representative target representations, 
thereby weakening the guidance provided by the teacher--student consistency constraint during representation learning. 
As the amount of observational information in teacher views increases, the model benefits from more stable and representative signals, leading to improved prediction accuracy.

In contrast, when $\alpha_{\text{T}}$ is further increased to 70\%, 
overall model performance degrades noticeably, with increased RMSE, reduced $R^2$, and larger prediction errors across all target metrics. 
This observation suggests that excessively high values of $\alpha_{\text{T}}$ substantially increase the informational disparity between teacher and student views, 
which in turn weakens the regularization effect induced by cross-view alignment and ultimately impairs model generalization.
Therefore, a teacher-view effective observation ratio of approximately 50\% provides a reasonable balance between information sufficiency and view diversity, 
and is adopted as the default setting in subsequent experiments.

\subsection{Effect of Multi-View Composition in the KD Framework}

In the current investigation, we examine how different multi-view compositions influence the prediction performance of TR metrics within the KD framework. 
Specifically, we fixed the observation rate of SMBG-T at $\alpha_{\mathrm{T}} = 50\%$ and SMBG-S at $\alpha_{\mathrm{S}} = 3\%$. Under this setting, we systematically vary the number of SMBG-T views ($N_\mathrm{T}$) and the number of SMBG-S views ($N_\mathrm{S}$) to examine how the balance between teacher guidance and student diversity affects the final TR metric estimation.
We design 7 different combinations of SMBG-T and SMBG-S view compositions, denoted as V0--V6. The specific configurations of each combination and the corresponding prediction performance are presented in Table~\ref{tab:view_count_sweep}. 

\begin{table*}[htbp]
\centering
\scriptsize
\caption{Impact of teacher and student view configurations on TR prediction performance.}
\label{tab:view_count_sweep}
\renewcommand{\arraystretch}{1.5}
\setlength{\tabcolsep}{5pt}
\begin{tabular}{|c|ccccccc|}
\hline
\text{Group} & $N_\mathrm{T}$ & $N_\mathrm{S}$ & \text{Overall RMSE} & \text{Overall $\text{R}^2$} & \text{TAR MAE} & \text{TIR MAE} & \text{TBR MAE} \\
\hline
V0 & 2 & 4 & 0.053 & 0.677 & 0.050 & 0.047 & 0.016 \\ \hline
V1 & 1 & 4 & 0.058 & 0.633 & 0.054 & 0.052 & 0.017 \\ \hline
V2 & 3 & 4 & 0.057 & 0.656 & 0.054 & 0.052 & 0.017 \\ \hline
V3 & 2 & 2 & 0.057 & 0.667 & 0.054 & 0.052 & 0.016 \\ \hline
V4 & 2 & 6 & 0.056 & 0.658 & 0.053 & 0.051 & 0.016 \\ \hline
V5 & 1 & 2 & 0.061 & 0.624 & 0.058 & 0.055 & 0.017 \\ \hline
V6 & 3 & 6 & 0.058 & 0.643 & 0.055 & 0.054 & 0.017 \\
\hline
\end{tabular}
\end{table*}

First, we examine the effect of varying $N_\mathrm{T}$ while keeping $N_\mathrm{S}$ fixed. This analysis is conducted using configurations V0, V1, and V2, where $N_\mathrm{T}$ ranges from 1 to 3, while $N_\mathrm{S}$ is fixed at 4.
Among these configurations, V0 achieves the best overall performance in terms of both RMSE and $R^2$, suggesting that this teacher--student view composition (2:4) more effectively supports knowledge distillation from the teacher to the student path. The results indicate that the level of diversity of the SMBG-T has a non-negligible impact on prediction performance.

When the diversity of teacher views is limited (V1), the supervision signal provided by the teacher path becomes relatively limited.
As a result, the corresponding representation targets are more susceptible to observation bias from sparse data.
However, when teacher representation is overly emphasized—such as in configuration V2 with three SMBG-T views—the TR prediction performance does not further improve and may even degrade. This is likely because increasing the number of teacher views can introduce correlation among them, leading to redundant representation targets.
In contrast, the teacher--student view composition adopted in V0 achieves a more balanced trade-off between representational stability and flexibility.
This configuration allows the teacher path to provide sufficiently stable targets, while preserving the student path’s ability to learn robust representations, leading to the best performance among the compared settings.


Under a fixed $N_\mathrm{T}$, we investigate the impact of varying $N_\mathrm{S}$ on prediction performance. This analysis is conducted using configurations V0, V3, and V4, where $N_\mathrm{T}$ is fixed at 2, and $N_\mathrm{S}$ ranges from 2 to 6.
The results show that V0 achieves the best overall performance in terms of both RMSE and $R^2$.
When $N_\mathrm{S}$ is small (V3), the diversity of sparse observation views available to the student encoder is significantly limited.
Such a narrow range of views is insufficient to capture the variability in SMBG sampling patterns, thereby restricting the model’s ability to learn robust and generalized representations for TR estimation.
When $N_\mathrm{S}$ becomes excessive (V4), the performance no longer improves. Although increasing the number of student views nominally expands the diversity of sparse observations, such expansion does not necessarily translate into proportional gains in effective information under a fixed sampling sparsity. Moreover, the student encoder is required to align representations across a larger set of highly sparse views, which increases the optimization difficulty of the self-distillation process.
Consequently, the student-view configuration adopted in V0 provides a favorable trade-off under the current setting. 


Finally, we examine how the absolute number of views affects prediction performance when the ratio between SMBG-T and SMBG-S views is held constant. Specifically, we consider configurations V0, V5, and V6, where the ratio of SMBG-T to SMBG-S views is fixed at 1:2. Under this controlled ratio, we vary the total number of views to assess whether increasing the overall view count leads to improved TR prediction. 

When both $N_\mathrm{T}$ and $N_\mathrm{S}$, as well as the total number of views, are small (V5), the overall performance degrades substantially. This is because the limited number of fused views provides insufficient informative representation and sparse observation coverage. As a result, the student encoder lacks access to sufficiently rich signals, hindering its ability to learn robust representations.
On the other hand, increasing both SMBG-T and SMBG-S views simultaneously to a large number (V6) does not lead to performance improvement. Instead, this configuration introduces an excessive number of distillation and alignment targets. These additional views do not necessarily provide new effective information and may instead exacerbate the difficulty of representation alignment and increase the optimization burden of the KD process.
In contrast, the configuration adopted in V0 strikes a more favorable balance in terms of absolute view scale. With a moderate number of teacher and student views, V0 provides sufficient informative representation and view diversity, while avoiding excessive redundancy and optimization complexity.

\subsection{Model Validation and Ablation Analysis of PACD-Net}

To systematically evaluate the contribution of each key component in the proposed PACD-Net framework, we conduct a comprehensive ablation study by varying the backbone architecture and self-supervised learning objectives. Note that all ablation experiments are performed under identical training configurations. 
Specifically, we evaluate three reduced model variants and compare them against the full PACD-Net model:
\begin{itemize}
    \item \textbf{CNN + Supervised Learning}: A baseline model using a CNN backbone trained solely with supervised loss on TR metrics.
    \item \textbf{Swin-CRB + Supervised Learning}: A stronger backbone using Swin-CRB, also trained with only supervised objectives.
    \item \textbf{Swin-CRB + KD}: A variant that incorporates KD, where teacher pseudo-SMBG with a high observation rate provide guidance to the student encoder, aiming to stabilize prediction through improved representation learning.
\end{itemize}

\begin{table*}[htbp]
\centering
\scriptsize
\caption{Model validation and ablation study of the PACD-Net framework.
Performance is evaluated using overall regression metrics and class-wise MAE for TAR, TIR, and TBR. MVC refers to multi-view contrastive learning.}
\label{tab:ablation_s2swin}
\renewcommand{\arraystretch}{1.5}
\setlength{\tabcolsep}{5pt}
\begin{tabular}{|l|ccccc|}
\hline
\text{Model Variant} 
& \text{RMSE}
& \text{$\text{R}^2$}
& \text{MAE$_{\text{TAR}}$} 
& \text{MAE$_{\text{TIR}}$}
& \text{MAE$_{\text{TBR}}$}\\
\hline
CNN Backbone w/ Supervised Regression 
& 0.084 & 0.456 & 0.085 & 0.081 & 0.017 \\ \hline
Swin-CRB w/ Supervised Regression 
& 0.068 & 0.539 & 0.064 & 0.061& 0.019 \\ \hline
Swin-CRB w/ KD-based Self-Distillation 
& 0.061 & 0.594 & 0.058 & 0.056 & 0.018 \\ \hline
PACD-Net (Swin-CRB w/ KD + MVC)
& 0.053 & 0.677 & 0.050 &0.047 & 0.016 \\
\hline
\end{tabular}
\end{table*}

\begin{figure}[htbp]
\centering
\caption{Scatter plots comparing predicted and ground truth glycemic metrics for different ablation variants.}
\label{fig:ablation_s2swin_scatter}
\begin{subfigure}[t]{\linewidth}
  \centering
  \includegraphics[width=0.88\linewidth]{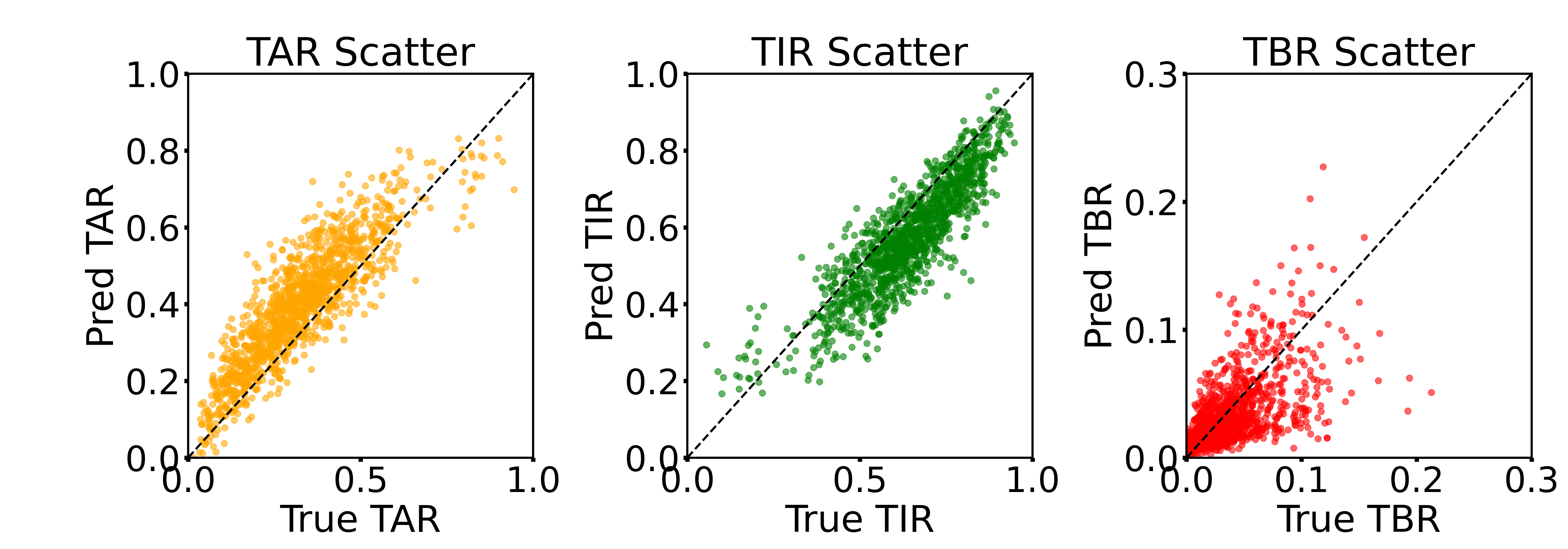}
  \caption{CNN backbone trained with supervised regression only.}
  \label{fig:ablation_cnn}
\end{subfigure}

\vspace{0.1cm}

\begin{subfigure}[t]{\linewidth}
  \centering
  \includegraphics[width=0.88\linewidth]{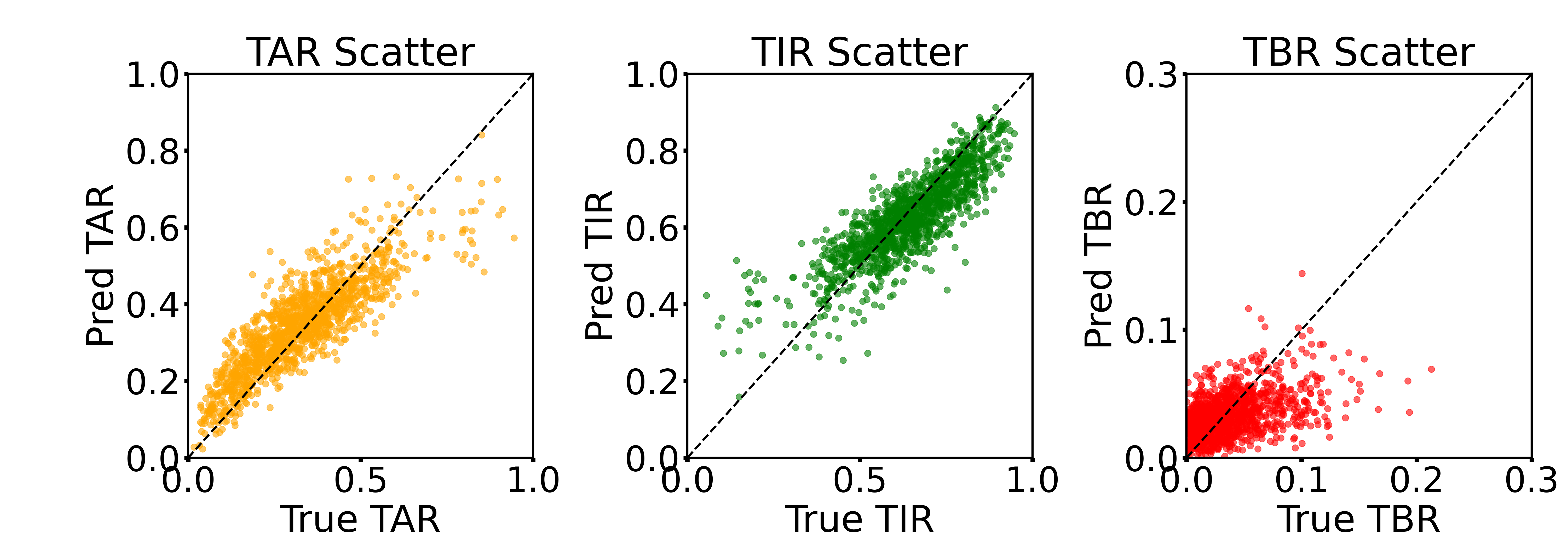} 
  \caption{Swin-CRB trained with supervised regression only.}
  \label{fig:ablation_swin_only}
\end{subfigure}

\vspace{0.1cm}

\begin{subfigure}[t]{\linewidth}
  \centering
  \includegraphics[width=0.88\linewidth]{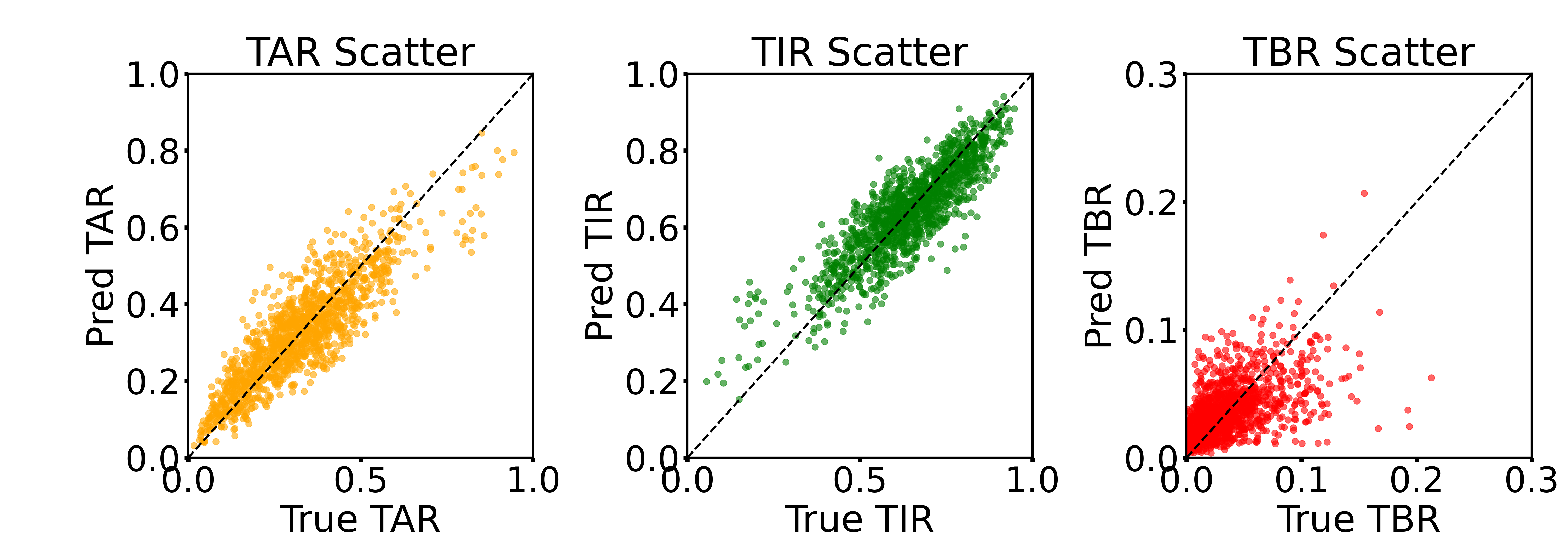} 
  \caption{Swin-CRB with KD-based self-distillation.}
  \label{fig:ablation_swin_KD}
\end{subfigure}

\vspace{0.1cm}

\begin{subfigure}[t]{\linewidth}
  \centering
  \includegraphics[width=0.88\linewidth]{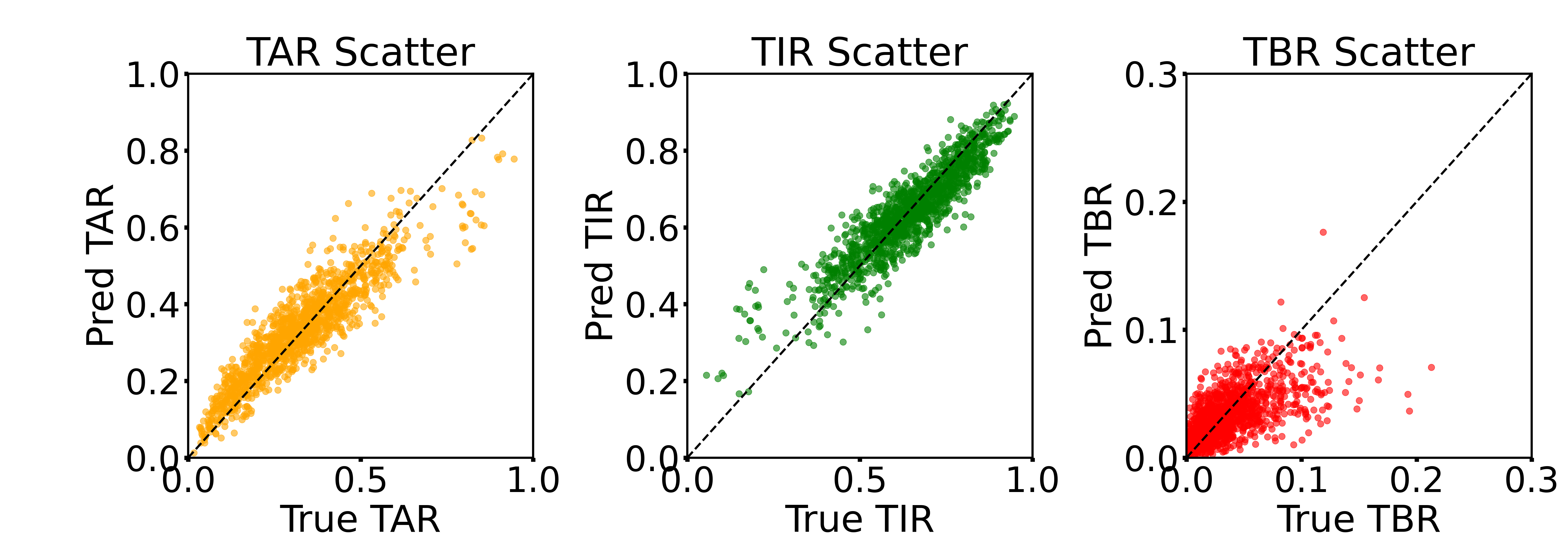} 
  \caption{PACD-Net model with joint KD and Multi-View Contrastive Learning.}
  \label{fig:ablation_PACD-Net}
\end{subfigure}

\end{figure}

The detailed quantitative results across all evaluation metrics are reported in Table~\ref{tab:ablation_s2swin}. In addition, scatter plots are provided in Fig.~\ref{fig:ablation_s2swin_scatter}, illustrating the predictions of different variants versus ground truth for TAR, TIR, and TBR at the sample level (each dot represents one sample).

As shown in Fig.~\ref{fig:ablation_s2swin_scatter}(a), the CNN-based pure supervised model exhibits significant systematic bias in the scatter plots, characterized by consistent overestimation of TAR and underestimation of TIR. In addition, the predicted values show a widely dispersed distribution, indicating limited predictive reliability. 
These patterns reflect the inherent limitations of CNN architectures, including a restricted receptive field and weaker capability in modeling long-range temporal dependencies, which hinder effective capture of multi-scale glycemic dynamics under sparse SMBG observations.

Compared with the CNN backbone, the proposed Swin-CRB backbone reduces systematic bias to some extent and produces a more concentrated distribution of predictions in Fig.~\ref{fig:ablation_s2swin_scatter}(b). Nevertheless, noticeable overestimation of TAR and underestimation of TIR remain evident in the scatter plots, and the distribution remains substantially more dispersed than that of the full PACD-Net model. 
These observations suggest that, under highly sparse and irregular SMBG observation settings, relying solely on limited supervised signals is insufficient to fully capture underlying glycemic dynamics. 
    
With the introduction of KD-based self-distillation, systematic bias is largely alleviated, and prediction consistency is visibly improved in the scatter plots (Fig.~\ref{fig:ablation_s2swin_scatter}(c)). 
This behavior indicates that KD significantly enhances representation consistency, mitigating the challenges posed by observation sparsity by enabling the extraction of richer patterns from teacher views.

In contrast, the full PACD-Net model shows further significant improvement (Fig.~\ref{fig:ablation_s2swin_scatter}(d)), with predictions exhibiting the tightest clustering along the diagonal among all variants. These results demonstrate that jointly incorporating KD and multi-view contrastive learning on top of the Swin-CRB backbone substantially enhances the model’s ability to estimate glycemic TR metrics under sparse SMBG observations. The design of PACD-Net is specifically tailored to address the challenges of high sparsity and irregular sampling patterns in SMBG data, enabling reliable estimation of TR metrics for assessing glycemic control status.

\subsection{Comparison Study with State-of-the-Art methods}

To comprehensively evaluate model performance, we compare the proposed PACD-Net framework with two state-of-the-art approaches: 

\begin{itemize}
    \item \textbf{SMBG No-Interpolation Baseline (No-Interp)}: This method directly computes the proportions of SMBG measurements that fall within clinically defined ranges corresponding to TAR, TIR, and TBR, without performing any temporal interpolation or signal reconstruction between successive observations. It represents the most straightforward use of raw SMBG data without introducing additional modeling assumptions and is commonly adopted in current clinical practice.
    \item \textbf{DPA-Net}: DPA-Net~\cite{lei2025dpa} is a supervised learning approach that jointly reconstructs CGM signals and predicts time-in-range (TR) metrics from sparse SMBG inputs. It represents a recent state-of-the-art method for translating SMBG data into TR metrics. 
\end{itemize}

\begin{table*}[htbp]
\centering
\scriptsize
\caption{Performance comparison of baseline and learning-based methods on the \textit{repbg} test set.
Results are reported as mean $\pm$ standard deviation over multiple runs.}
\label{tab:baseline_overall}
\renewcommand{\arraystretch}{1.5}
\setlength{\tabcolsep}{4pt}
\begin{tabular}{|l|ccccc|}
\hline
Model & RMSE & Overall $\text{R}^2$ & TAR MAE & TIR MAE & TBR MAE \\
\hline
SMBG Baseline
& $0.099$
& $0.864$
& $0.093$
& $0.097$
& $0.022$ \\ \hline

DPA-Net
& $0.061 \pm 0.011$
& $0.615 \pm 0.074$
& $0.058 \pm 0.010$
& $0.059 \pm 0.012$
& $0.018 \pm 0.006$ \\ \hline

PACD-Net
& $0.053 \pm 0.003$
& $0.677 \pm 0.025$
& $0.050 \pm 0.004$
& $0.047 \pm 0.003$
& $0.016 \pm 0.002$ \\
\hline
\end{tabular}
\end{table*}

The performance metrics are summarized in Table~\ref{tab:baseline_overall}. The No-Interp baseline yields the highest RMSE and MAE across all three TR metrics.
It is worth noting that the deep learning-based methods, including DPA-Net and PACD-Net, are evaluated over multiple runs with different random network initializations to estimate performance variability and ensure the reliability and repeatability of the predictions.
DPA-Net demonstrates improved performance across all TR evaluation metrics compared with No-Interp, indicating that leveraging population-level data can enhance TR metric prediction for unseen SMBG samples.
In contrast, the proposed PACD-Net achieves the best overall performance, with the lowest average RMSE and MAE, as well as the highest $\text{R}^2$ values across all runs compared with the other methods. Moreover, PACD-Net exhibits significantly smaller standard deviations, indicating more stable optimization dynamics and stronger robustness compared to conventional supervised learning approaches.

\begin{figure}[htbp]
\centering
\caption{Predicted vs. ground-truth AGP glycemic metric scatter plots for baseline, DPA model, and our method.}
\label{fig:scatter-compare-vert}

\begin{subfigure}[t]{\linewidth}
  \centering
  \includegraphics[width=\linewidth]{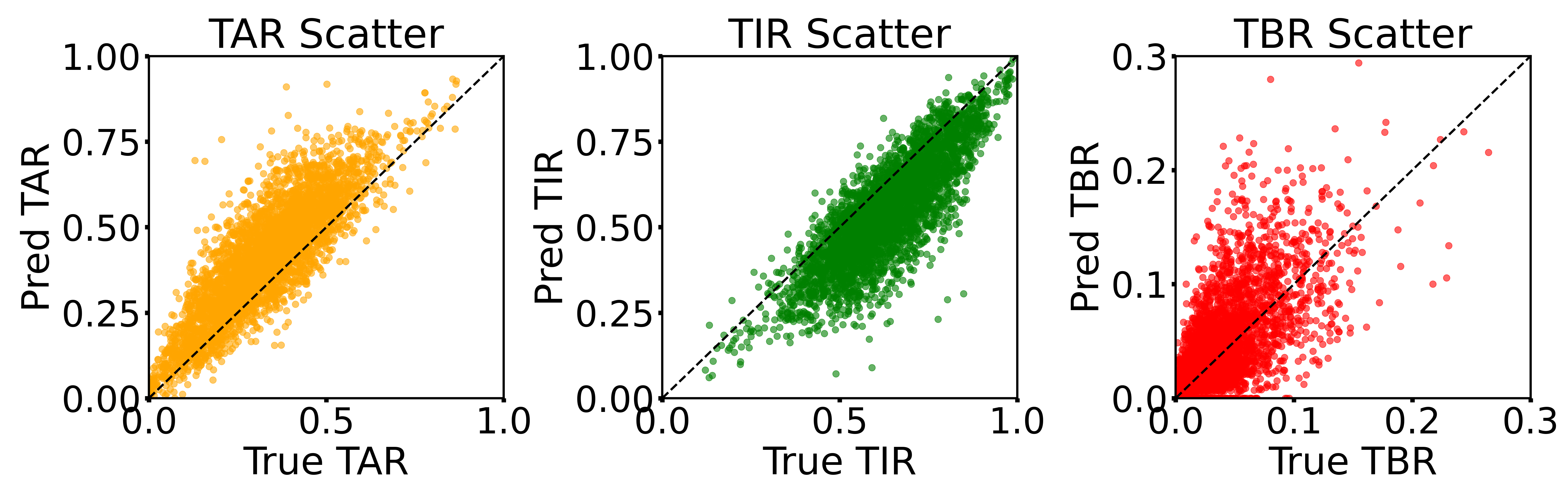}
  \caption{SMBG baseline (No-Interp)}
  \label{fig:baseline_scatter}
\end{subfigure}

\vspace{0.1cm}

\begin{subfigure}[t]{\linewidth}
  \centering
  \includegraphics[width=\linewidth]{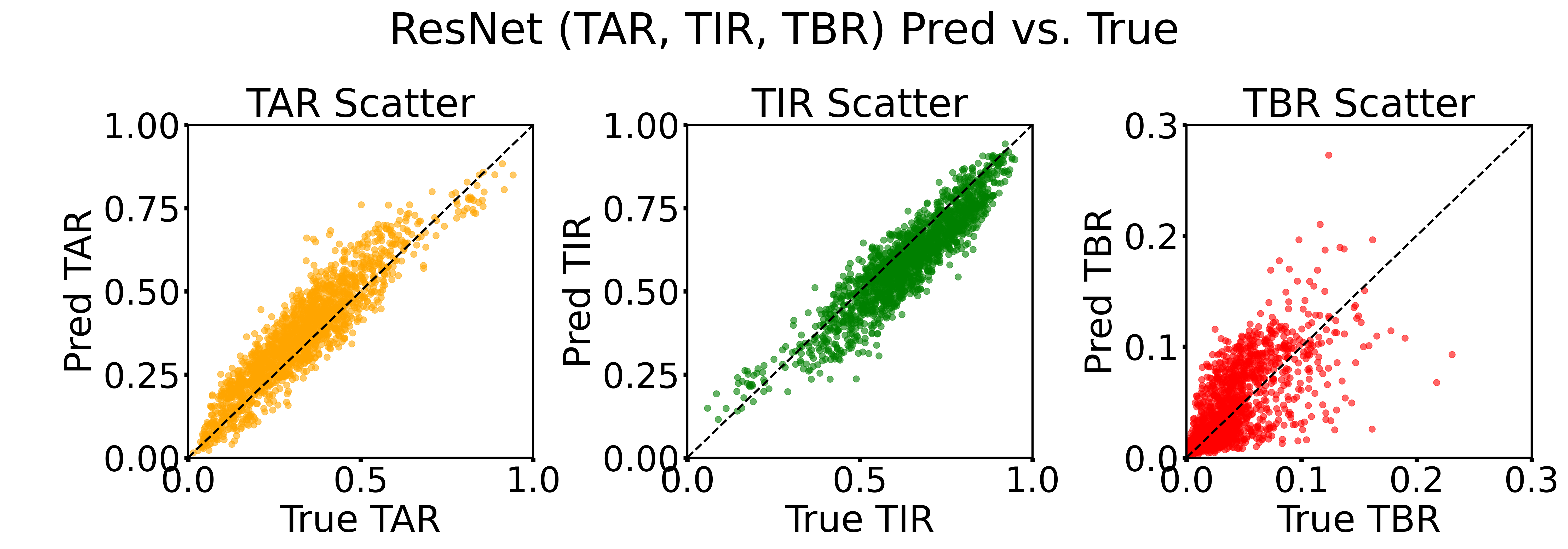} 
  \caption{DPA-Net}
  \label{fig:dpa_scatter}
\end{subfigure}

\vspace{0.1cm}

\begin{subfigure}[t]{\linewidth}
  \centering
  \includegraphics[width=\linewidth]{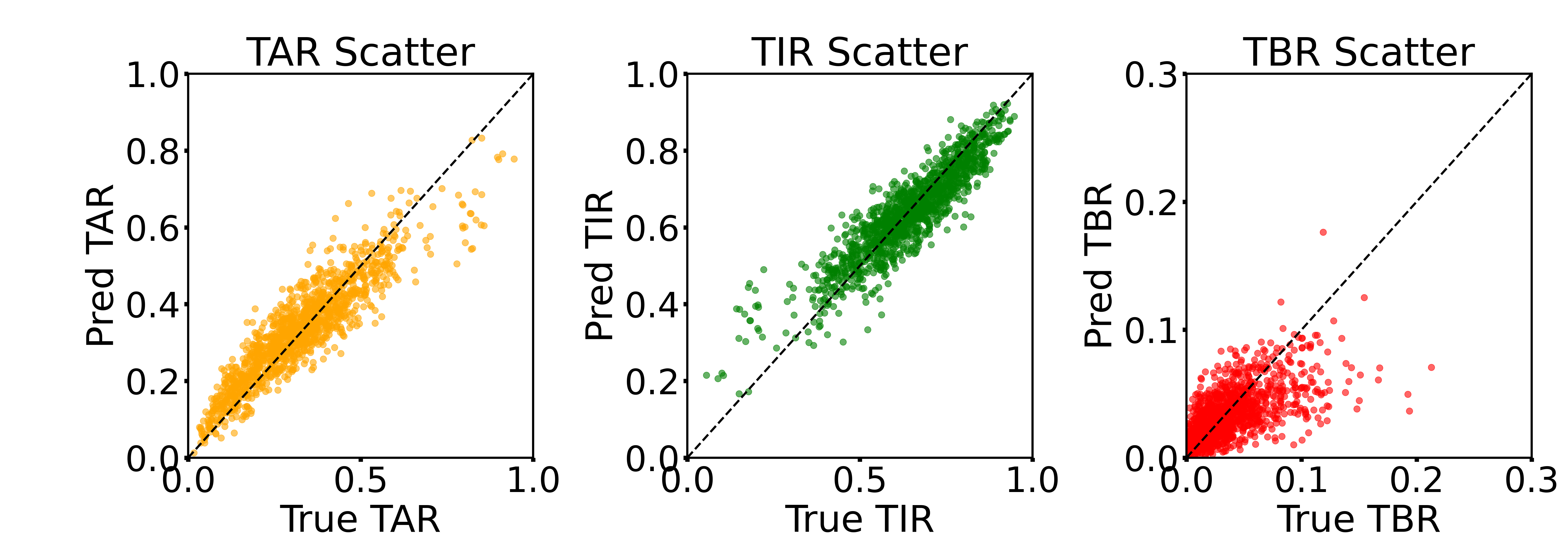} 
  \caption{ PACD-Net}
  \label{fig:s2_scatter}
\end{subfigure}

\end{figure}

Fig.~\ref{fig:scatter-compare-vert} illustrates the scatter plots of predicted versus ground-truth TR metrics for different methods, providing a visualization of their predictive behaviors.
Fig.~\ref{fig:baseline_scatter} shows the TR predictions generated by the SMBG No-Interp baseline. 
A pronounced systematic bias is observed, characterized by overestimation of TAR and underestimation of TIR, suggesting that the SMBG baseline fails to effectively correct the bias introduced by sparse SMBG sampling.
Fig.~\ref{fig:dpa_scatter} shows that DPA-Net improves the alignment between predictions and ground truth compared to the SMBG baseline, with a more concentrated point distribution overall. Nevertheless, noticeable dispersion remains, particularly in the TBR region where samples are relatively sparse. Although the systematic bias is partially alleviated relative to No-Interp, the tendency to overestimate TAR and underestimate TIR is still evident from the scatter plots, indicating that the prediction stability of DPA-Net remains limited across different experimental conditions.
In contrast, Fig.~\ref{fig:s2_scatter} demonstrates that PACD-Net produces more compact scatter distributions across all three glycemic metrics, with predictions being more consistently aligned along the diagonal. The systematic bias is reduced, and the overall prediction variance is smaller (Table~\ref{tab:baseline_overall}). These observations indicate that the PACD-Net framework, which integrates multi-view self-supervised representation learning with an improved supervised prediction branch, achieves more stable generalization behavior and superior overall predictive performance under sparse SMBG settings.

\section{Conclusion}

In this study, we proposed PACD-Net, a self-supervised framework for estimating glycemic control status by predicting TR metrics—including TAR, TIR, and TBR—from sparse SMBG data, thereby enabling accurate clinical assessment under limited observation conditions. The proposed framework adopts a modified Swin Transformer as the backbone encoder, augmented with convolutional residual blocks (CRB), combined with knowledge distillation and multi-view contrastive learning to facilitate robust representation learning. 
In particular, the integration of knowledge distillation enhances representation consistency by aligning the student encoder with teacher encoder that contain richer patterns, while multi-view learning improves generalization under diverse sampling behaviors. These findings highlight the effectiveness of the proposed framework in extracting meaningful representations from sparse and irregular observations. Extensive experimental results demonstrate that PACD-Net consistently outperforms existing approaches, achieving superior accuracy and stability across multiple evaluation metrics. 
Overall, PACD-Net provides a reliable and robust solution for estimating glycemic control metrics from SMBG data, offering a practical pathway toward more accessible and cost-effective diabetes monitoring in real-world settings. Beyond this application, the proposed framework also offers a general paradigm for representation learning under sparse and irregular observations, with potential to benefit other sparse data analysis settings.

\bibliography{refs} 
\end{document}